\pdfoutput=1

\documentclass[11pt]{article}
\usepackage{graphicx}

\usepackage[]{acl}

\usepackage{times}
\usepackage{latexsym}
\usepackage{amssymb}
\usepackage{amsmath}
\usepackage{mathtools, nccmath}
\usepackage{tabularx}         
\usepackage{booktabs}         
\usepackage{multirow,multicol}         
\usepackage{cuted}

\usepackage[T1]{fontenc}

\usepackage[utf8]{inputenc}

\usepackage{microtype}

\title{Traditional Readability Formulas Compared for English}

\author{Bruce W. Lee$^{1,2}$ \\ University of Pennsylvania$^{1}$ \\ Pennsylvania, USA \\ brucelws@seas.upenn.edu
         \And
         Jason Hyung-Jong Lee$^{2}$ \\ LXPER AI Research (LAIR)$^{2}$  \\ Seoul, South Korea \\ jasonlee@lxper.com}

\begin{document}
\maketitle
\begin{abstract}
Traditional English readability formulas, or equations, were largely developed in the 20th century. Nonetheless, many researchers still rely on them for various NLP applications. This phenomenon is presumably due to the convenience and straightforwardness of readability formulas. In this work, we contribute to the NLP community by 1. introducing New English Readability Formula (NERF), 2. recalibrating the coefficients of ``old'' readability formulas (Flesch-Kincaid Grade Level, Fog Index, SMOG Index, Coleman-Liau Index, and Automated Readability Index), 3. evaluating the readability formulas, for use in text simplification studies and medical texts, and 4. developing a Python-based program for the wide application to various NLP projects.
\end{abstract}

\section{Introduction}
Readability Assessment (RA) quantitatively measures the ease of understanding or comprehension of any written text \citep{feng:2010, klare:2000}. Understanding text readability, or difficulty, is essential for research on any originated, studied, or shared ideas \citep{thompson:2014}. Such inherent property leads to RA's close applications to various areas of healthcare \citep{wu:2013}, education \citep{dennis:2018}, communication \citep{zhou:2017}, and Natural Language Processing (NLP), such as text simplification \citep{aluisio:2010}.

Machine learning (ML) or transformer-based methods have been reasonably successful in RA. The RoBERTa-RF-T1 model by \citet{lee:2021} achieves a $99\%$ classification accuracy on OneStopEnglish dataset \citep{onestop} and a BERT-based ReadNet model from \citet{meng:2020t} achieves about $92\%$ accuracy on WeeBit dataset \citep{weebit}. However, ``traditional readability formulas'' still seem to be actively used throughout the research published in popular NLP venues like ACL or EMNLP \citep{uchendu:2020,shardlow:2019,scarton:2018,schwartz:2017,xu:2016}. The tendency to opt for traditional readability formulas is likely due their convenience and straightforwardness. 

In this work, we hope to assist the NLP community by recalibrating five traditional readability formulas -- originally developed upon 20th-century military or technical documents. The formulas are adjusted for the modern, standard U.S. education curriculum. We utilize the appendix B (Text Exemplars and Sample Performance Tasks) dataset, provided by the U.S. Common Core State Standards\footnote{corestandards.org}. Then, we evaluate the performances and applications of these formulas. Lastly, we develop a Python-based program for convenient application of the recalibrated versions.

But traditional readability formulas lack wide linguistic coverage \citep{feng:2010}. Therefore, we create a \textit{new formula} that is mainly motivated by lexico-semantic and syntactic linguistic branches, as identified by \citet{thompson:2014}. From each, we search for the representative features. The resulting formula is named the New English Readability Formula, or simply \textbf{NERF}, and it aims to give the most generally and commonly accepted approach to calculating English readability.

To sum up, we make the contributions below. The related public resources are in appendix A.

\noindent \textbf{1.} We recalibrate five traditional readability formulas to show higher prediction accuracy on modern texts in the U.S. curriculum.

\noindent \textbf{2.} We develop NERF, a generalized and easy-to-use readability assessment formula. 

\noindent \textbf{3.} We evaluate and cross-compare six readability formulas on several datasets. These datasets are carefully selected to collectively represent the diverse audiences, education curricula, and reading levels.

\noindent \textbf{4.} We develop <Anonymous>, a fast open-source readability assessment software based on Python.

\section{Related Work}
The earliest attempt to "calculate" text readability was by \citet{lively:1923}, in response to their practical problem of selecting science textbooks for high school students \citep{dubay:2004}. In the consecutive years, many well-known readability formulas were developed, including Flesch Kincaid Grade Level \citep{kincaid:1975}, Gunning Fog Count (or Index) \citep{gunning:1952}, SMOG Index \citep{smog:1969}, Coleman-Liau Index \citep{coleman:1975}, and Automated Readability Index \citep{smith:1967}. 

These formulas are mostly linear models with two or three variables, largely based on superficial properties concerning words or sentences \citep{feng:2010}. Hence, they can easily combine with other systems with less burden of a large trained model \citep{xu:2016}. Such property also proved helpful in research fields outside computational linguistics, with some applications directly related to the public medical knowledge -- measuring the difficulty of a patient material \citep{gaeta:2021, van:2021, bange:2019, haller:2019, hansberry:2018, kiwanuka:2017}.  

\section{Datasets}
\subsection{Common Core - Appendix B (CCB)}
We use the CCB corpus to calibrate formulas. The article excerpts included in CCB are divided into the categories of story, poetry, informational text, and drama. For the simplification of our approach, we limit our research to story-type texts. This left us with only 69 items to train with. But those are directly from the U.S. Common Core Standards. Hence, we assume with confidence that the item classification is generally agreeable in the U.S. 

\begin{table}
\centering
\resizebox{0.5\textwidth}{!}{%
\begin{tabular}{lcccccc}
\cmidrule(lr){1-7}
\textbf{Properties}    & \textbf{CCB} & \textbf{WBT}  & \textbf{CAM} & \textbf{CKC} & \textbf{OSE} & \textbf{NSL}\\ 
\cmidrule(lr){1-1}\cmidrule(lr){2-2}\cmidrule(lr){3-3}\cmidrule(lr){4-4}\cmidrule(lr){5-5}\cmidrule(lr){6-6}\cmidrule(lr){7-7}
audience                      & Ntve        & Ntve    & ESL          & ESL    & ESL    & Ntve    \\
grade                         & K1-12       & K2-10   & A2-C2        & S7-12  & N/A    & N/A  \\
curriculum?                   & Yes         & No      & Yes          & Yes    & No     & No    \\
balanced?                     & No          & Yes     & Yes          & No     & Yes    & No     \\
\#class                       & 6           & 5       & 5            & 6      & 3      & 5      \\
\#item/class                  & 11.5        & 625     & 60.0         & 554    & 189    & 2125    \\
\#word/item                   & 362         & 213     & 508          & 117    & 669    & 752    \\
\#sent/item                   & 25.8        & 17.0    & 28.4         & 54.0   & 35.6   & 50.9   \\
\cmidrule(lr){1-7}
\end{tabular}
}
\caption{\label{Table 1} Modified data. These stats are based on respective original versions. S: S.Korea Grade, Ntve: Native}
\vspace{-4mm}
\end{table}

CCB is the only dataset that we use in the calibration of our formulas. All below datasets are mainly for feature selection purposes only. 

\subsection{WeeBit (WBT)}
WBT, the largest native dataset available in RA, contains articles targeted at readers of different age groups from the Weekly Reader magazine and the BBC-Bitesize website. In table 1, we translate those age groups into U.S. schools' K-* format. We downsample to $625 \frac{\text{item}}{\text{class}}$ as per common practice.

\subsection{Cambridge English (CAM)}
CAM \citep{xia:2016} classifies 300 items in the Common European Framework of Reference (CEFR) \citep{council:2001}. The passages are from the past reading tasks in the five main suites Cambridge English Exams (KET, PET, FCE, CAE, CPE), targeted at learners at A2–C2 levels of CEFR.

\subsection{Corpus of the Korean ELT (English Lang. Train.) Curriculum (CKC)}
CKC \citep{lee:2020a,lee:2020} is less-explored. It developed upon the reading passages appearing in the Korean English education curriculum. These passages' classifications are from official sources from the Korean Ministry. CKC represents a non-native country's official ESL education curriculum.

\subsection{OneStopEnglish (OSE)}
OSE is a recently developed dataset in RA. It aims at ESL (English as Second Language) learners and consists of three paraphrased versions of an article from The Guardian Newspaper. Along with the original OSE dataset, we created a paired version (OSE-Pair). This variation has 189 items and each item has advanced-intermediate-elementary pairs.

In addition, OSE-Sent is a sentence-paired version of OSE. The dataset consists of three parts: adv-ele (1674 pairs), adv-int (2166), int-ele (2154).

\subsection{Newsela (NSL)}
NSL \citep{xu:2015} is a dataset particularly developed for text simplification studies. The dataset consists of 1,130 articles, with each item re-written 4 times for children at different grade levels. We create a paired version (NSL-Pair) (2125 pairs). 

\subsection{ASSET}
ASSET \citep{asset} is a paired sentence dataset. The dataset consists of 360 sentences, with each item simplified 10 times.

\section{Recalibration}
\subsection{Choosing Traditional Read. Formulas}
We start by recalibrating five readability formulas. We considered \citet{zhou:2017} and the number of Google Scholar citations to sort out the most popular traditional readability formulas. Further, to make a fair performance comparison with our adjusted variations, we choose the formulas originally intended to output U.S. school grades but are based on 20th-century texts and test subjects.

Flesh-Kincaid Grade Level (FKGL) is primarily developed for U.S. Navy personnel. The readability level of 18 passages from Navy technical training manuals was calculated. The criterion was that $50\%$ of subjects with reading abilities at the specific level had to score $\geq 35\%$ on a cloze test for a text item to be classified as the specific reading level. Responses from 531 Navy personnel were used.

\begin{equation*}
\text{FKGL} = a \cdot \frac{\text{\#word}}{\text{\#sent}} + b \cdot \frac{\text{\#syllable}}{\text{\#word}} + c
\end{equation*}
where sent is sentence, and \# refers to "count of." 

The genius of Gunning Fog Index (FOGI) is the idea that word difficulty highly correlates with the number of syllables. Such a conclusion was deduced upon the inspection of Dale's list of easy words \citep{zhou:2017, dale:1948}. However, the shortcoming of FOGI is the over-generalization that "all" words with more than two syllables are difficult. Indeed, "banana" is quite an easy word.

\begin{equation*}
\text{FOGI} = a \cdot (\frac{\text{\#word}}{\text{\#sent}} + b \cdot \frac{\text{\#difficult word}}{\text{\#word}}) + c
\end{equation*}

Simple Measure of Gobbledygook (SMOG) Index, known for its simplicity, resembles FOGI in that both use the number of syllables to classify a word's difficulty. But SMOG sets its criterion a little high to more than three syllables per word. Additionally, SMOG incorporates a square root approach instead of a linear regression model. 

\begin{equation*}
\text{SMOG} = a \cdot \sqrt{b \cdot \frac{\text{\#polysyllable word}}{\text{\#sent}}} + c
\end{equation*}

Coleman-Liau Index (COLE) is more of a lesser-used variation among the five. But we could still find multiple studies outside computational linguistics that still partly depend on COLE \citep{kue:2021,szmuda:2020, joseph:2020, powell:2020}. The novelty of COLE is that it calculates readability without counting syllables, which was viewed as a time-consuming approach.

\begin{equation*}
\text{COLE} = a \cdot 100 \cdot \frac{\text{\#letter}}{\text{\#word}} + b \cdot 100 \cdot \frac{\text{\#sent}}{\text{\#word}} + c
\end{equation*}

Automated Readability Index (AUTO) is developed for U.S. Air Force to handle more technical documents than textbooks. Like COLE, AUTO relies on the number of letters per word, instead of the more commonly-used syllables per word. Another quirk is that non-integer scores are all rounded up.

\begin{equation*}
\text{AUTO} = a \cdot \frac{\text{\#letter}}{\text{\#word}} + b \cdot \frac{\text{\#word}}{\text{\#sent}} + c
\end{equation*}

\subsection{Recalibration \& Performance}
\subsubsection{Traditional Formulas, Other Text Types}
We only recalibrate formulas on the CCB dataset. As stated in section 2.1, we limit to CCB's story-type items. In a preliminary investigation, we obtained low r2 scores ($<0.3$, before and after recalibration) between the traditional readability formulas and poetry, informational text, and drama.

\subsubsection{Details on Recalibration}
We started with a large feature extraction software, LingFeat \citep{lee:2021} and expanded it to include more necessary features. From CCB texts, we extracted the surface-level features in traditional readability formulas (i.e. $\frac{\text{\#letter}}{\text{\#word}}$, $\frac{\text{\#word}}{\text{\#sent}}$, $\frac{\text{\#syllable}}{\text{\#word}}$) and put them in a dataframe. 

CCB has 6 readability classes, but they are in the forms of range: K1, K2-3, K4-5, K6-8, K9-10, K11, and CCR (college and above). During calibration and evaluation, we estimated readability classes to K1, K2.5, K4.5, K7, K9.5, or K12 to model the general trend of CCB.

Using the class estimations as true labels and the created dataframe as features, we ran an optimization function to calculate the best coefficients (a, b, c in \textbf{\S4.1}). We used non-linear least squares in fitting functions \citep{SciPy}. Additional details are available in appendix B. 

\subsubsection{Coefficients \& Performances}
\begin{table}[t]
\centering
\resizebox{0.5\textwidth}{!}{%
\begin{tabular}{lccccc}
\cmidrule(lr){1-6}
\textbf{a) Coef.s}    &\textbf{FKGL}&\textbf{FOGI}  &\textbf{SMOG} &\textbf{COLE} &\textbf{AUTO}\\ 
\cmidrule(lr){1-1}\cmidrule(lr){2-2}\cmidrule(lr){3-3}\cmidrule(lr){4-4}\cmidrule(lr){5-5}\cmidrule(lr){6-6}
original-a                    & 0.390    & 0.4000            & 1.043           & 0.05880        & 4.710  \\
\textbf{adjusted-a}     &\textbf{0.1014} & \textbf{0.1229}  & \textbf{2.694 } &\textbf{0.03993}& \textbf{6.000}  \\
original-b                    & 11.80    & 100.0            & 30.00           & -0.2960        & 0.5000  \\
\textbf{adjusted-b}     &\textbf{20.89}  & \textbf{415.7}   & \textbf{8.815 } &\textbf{-0.4976}& \textbf{0.1035}  \\
original-c                    & -15.59   & 0.0000           & 3.129           & -15.80         & -21.43  \\
\textbf{adjusted-c}     &\textbf{-21.94} & \textbf{1.866}   & \textbf{3.367 } &\textbf{-5.747} & \textbf{-19.61}  \\
\cmidrule(lr){1-6}
\textbf{b) Perf.} &\textbf{FKGL}&\textbf{FOGI}  &\textbf{SMOG} &\textbf{COLE} &\textbf{AUTO}\\ 
\cmidrule(lr){1-1}\cmidrule(lr){2-2}\cmidrule(lr){3-3}\cmidrule(lr){4-4}\cmidrule(lr){5-5}\cmidrule(lr){6-6}
r2 score                   & -0.03835   & -0.3905           & 0.1613           & 0.4341         & -0.5283  \\
\textbf{r2 score}     &\textbf{0.4423} & \textbf{0.4072}   & \textbf{0.3192} &\textbf{0.4830} & \textbf{0.4263}  \\
Pearson r                  & 0.5698     & 0.5757            & 0.5649           & 0.6800         & 0.5684  \\
\textbf{Pearson r}     &\textbf{0.6651} & \textbf{0.6381}   & \textbf{0.5649} &\textbf{0.6949} & \textbf{0.6529}  \\
\cmidrule(lr){1-6}
\end{tabular}
}
\caption{\label{Table 2} a) Original \& adjusted coefficients. b) Perform-ance on CCB. Measured on U.S. Standard Curriculum's K-* Output. Bold refers to our new, adjusted versions.}
\vspace{-4mm}
\end{table}

 Table 2-a shows the original coefficients and the adjusted variations, rounded up to match significant figures. The adjusted traditional readability formulas can be obtained by simply plugging in these values to the formulas in section 4.1.

\section{The New English Readability Formula}
\subsection{Criteria}
Considering the value of traditional readability formulas as essentially the generalized definition of readability for the non-experts (section 1), what really matters is the included features. The coefficients (or weights) can be recalibrated anytime to fit a specific use. Therefore, it is important to first identify handcrafted linguistic features that universally affect readability. Additionally, to ensure breadth and usability, we set the following guides:

\noindent \textbf{1.} We avoid surface-level features that lack linguistic value \citep{feng:2010}. They include $\frac{\text{\#letter}}{\text{\#word}}$.

\noindent \textbf{2.} We include at most one linguistic feature from each linguistic subgroup. We use the classifications from \citet{lee:2021, thompson:2014}.

\noindent \textbf{3.} We stick to a simplistic linear equation format.

\begin{table*}[t]
\centering
\resizebox{\textwidth}{!}{%
\begin{tabular}{
c@{\hspace{0.8ex}}
l@{\hspace{0.8ex}}
l@{\hspace{0.8ex}}
l@{\hspace{0.8ex}}
l@{\hspace{0.8ex}}
|c@{\hspace{0.8ex}}
c@{\hspace{0.8ex}}
|c@{\hspace{0.8ex}}
c@{\hspace{0.8ex}}
|c@{\hspace{0.8ex}}
c@{\hspace{0.8ex}}
|c@{\hspace{0.8ex}}
c@{\hspace{0.8ex}}
|c@{\hspace{0.8ex}}
c@{\hspace{0.8ex}}}
\cmidrule(lr){1-15}
\multicolumn{5}{c}{\textbf{Feature}} & \multicolumn{2}{c}{\textbf{CCB}} & \multicolumn{2}{c}{\textbf{WBT}} & \multicolumn{2}{c}{\textbf{CAM}} & \multicolumn{2}{c}{\textbf{CKC}} & \multicolumn{2}{c}{\textbf{OSE}} \\
\cmidrule(lr){1-5}\cmidrule(lr){6-7}\cmidrule(lr){8-9}\cmidrule(lr){10-11}\cmidrule(lr){12-13}\cmidrule(lr){14-15}
Score & Branch & Subgroup & LingFeat Code & Brief Explanation          &   r  &  rk &  r &  rk &  r  &  rk & r  &  rk& r  &   rk  \\
\cmidrule(lr){1-1}\cmidrule(lr){2-2}\cmidrule(lr){3-3}\cmidrule(lr){4-4}\cmidrule(lr){5-5}\cmidrule(lr){6-6}\cmidrule(lr){7-7}\cmidrule(lr){8-8}\cmidrule(lr){9-9}\cmidrule(lr){10-10}\cmidrule(lr){11-11}\cmidrule(lr){12-12}\cmidrule(lr){13-13}\cmidrule(lr){14-14}\cmidrule(lr){15-15}
43& LxSem  & Psycholinguistic &as\_AAKuL\_C& Kuperman Lemma AoA per Sent      &0.540&25 &0.505&1  &0.722&42 &0.711&4    &0.601&25       \\

43& LxSem  & Psycholinguistic &as\_AAKuW\_C& Kuperman Word AoA per Sent      &0.537&28 &0.503&2  &0.722&43 &0.711&6    &0.602&24       \\

40& LxSem  & Psycholinguistic &at\_AAKuW\_C& Kuperman Word AoA per Word       &0.703&5 &0.308&36  &0.784&20 &0.643&21   &0.455&66       \\

40& Synta  & Tree Structure   &as\_TreeH\_C& Tree Height per Sent               &0.550&21 &0.341&30 &0.686&51 &0.699&9    &0.541&44      \\

40& Synta  & Part-of-Speech   &as\_ContW\_C& \# Content Words per Sent           &0.534&29 &0.453&13 &0.667&56 &0.688&14   &0.544&43       \\

39& LxSem  & Psycholinguistic &at\_AAKuL\_C& Kuperman Lemma AoA per Word      &0.723&4  &0.323&35 &0.785&19 &0.650&20   &0.453&67       \\

39& Synta  & Phrasal          &as\_NoPhr\_C& \# Noun Phrases per Sent           &0.550&20 &0.406&25 &0.660&58 &0.673&18   &0.582&35       \\

39& Synta  & Phrasal          &to\_PrPhr\_C& Total \# Prepositional Phrases   &0.470&47 &0.189&58 &0.808&11 &0.580&36   &0.729&3       \\

39& Synta  & Part-of-Speech   &as\_FuncW\_C& \# Function Words per Sent      &0.468&48 &0.471&8  &0.662&57 &0.673&17   &0.614&19       \\

38& LxSem  & Psycholinguistic &to\_AAKuL\_C& Total Sum Kuperman Lemma AoA       &0.428&71 &0.189&59 &0.835&3  &0.627&22   &0.716&5       \\

38& LxSem  & Psycholinguistic &to\_AAKuW\_C& Total Sum Kuperman Word AoA       &0.427&72 &0.189&60 &0.835&4  &0.625&23   &0.715&6       \\

36& Synta  & Phrasal          &as\_PrPhr\_C& \# Prepositional Phrases per Sent  &0.513&35 &0.417&23 &0.607&70 &0.608&28   &0.590&34      \\

36& LxSem  & Word Familiarity &as\_SbL1C\_C& SubtlexUS Lg10CD Value per Sent  &0.467&49 &0.430&20 &0.612&69 &0.699&10   &0.533&45       \\

35& LxSem  & Type Token Ratio & CorrTTR\_S & Corrected Type Token Ratio     &0.745&1  &0.006&228 &0.846&1  &0.445&65   &0.692&7       \\

35& LxSem  & Word Familiarity &as\_SbL1W\_C& SubtlexUS Lg10WF Value per Sent  &0.462&52 &0.437&19 &0.605&71 &0.693&12   &0.523&48       \\

\cmidrule(lr){1-15}
\end{tabular}
}
\caption{\label{Table 3} Top 15 (score $\geq$ 35) handcrafted linguistic features under Approach A. r: Pearson's correlation between the feature and the dataset. rk: the feature's correlation ranking on the specific dataset.  Full version in appendix D.}
\end{table*}

\begin{table*}[t]
\centering
\resizebox{\textwidth}{!}{%
\begin{tabular}{
c@{\hspace{0.8ex}}
l@{\hspace{0.8ex}}
l@{\hspace{0.8ex}}
l@{\hspace{0.8ex}}
l@{\hspace{0.8ex}}
|c@{\hspace{0.8ex}}
c@{\hspace{0.8ex}}
|c@{\hspace{0.8ex}}
c@{\hspace{0.8ex}}
|c@{\hspace{0.8ex}}
c@{\hspace{0.8ex}}
|c@{\hspace{0.8ex}}
c@{\hspace{0.8ex}}
|c@{\hspace{0.8ex}}
c@{\hspace{0.8ex}}}
\cmidrule(lr){1-15}
\multicolumn{5}{c}{\textbf{Feature}} & \multicolumn{2}{c}{\textbf{CCB}} & \multicolumn{2}{c}{\textbf{WBT}} & \multicolumn{2}{c}{\textbf{CAM}} & \multicolumn{2}{c}{\textbf{CKC}} & \multicolumn{2}{c}{\textbf{OSE}} \\
\cmidrule(lr){1-5}\cmidrule(lr){6-7}\cmidrule(lr){8-9}\cmidrule(lr){10-11}\cmidrule(lr){12-13}\cmidrule(lr){14-15}
Score & Branch & Subgroup & LingFeat Code & Brief Explanation          &   r  &  rk &  r &  rk &  r  &  rk & r  &  rk& r  &   rk  \\
\cmidrule(lr){1-1}\cmidrule(lr){2-2}\cmidrule(lr){3-3}\cmidrule(lr){4-4}\cmidrule(lr){5-5}\cmidrule(lr){6-6}\cmidrule(lr){7-7}\cmidrule(lr){8-8}\cmidrule(lr){9-9}\cmidrule(lr){10-10}\cmidrule(lr){11-11}\cmidrule(lr){12-12}\cmidrule(lr){13-13}\cmidrule(lr){14-14}\cmidrule(lr){15-15}
35& LxSem  & Psycholinguistic &as\_AAKuL\_C& Kuperman Lemma AoA per Sent      &0.540&25 &0.505&1  &0.722&42 &0.711&4    &0.601&25       \\

35& LxSem  & Psycholinguistic &as\_AAKuW\_C& Kuperman Word AoA per Sent      &0.537&28 &0.503&2  &0.722&43 &0.711&6    &0.602&24       \\

32& LxSem  & Psycholinguistic &at\_AAKuL\_C& Kuperman Lemma AoA per Word      &0.723&2 &0.323&35  &0.785&42 &0.650&22   &0.453&67       \\

32& LxSem  & Psycholinguistic &at\_AAKuW\_C& Kuperman Word AoA per Word       &0.703&5 &0.308&36  &0.784&20 &0.643&21   &0.455&66       \\

31& Synta  & Phrasal          &as\_NoPhr\_C& \# Noun Phrases per Sent           &0.550&20 &0.406&25 &0.660&58 &0.673&18   &0.582&35       \\

31& Synta  & Part-of-Speech   &as\_ContW\_C& \# Content Words per Sent           &0.534&29 &0.453&13 &0.667&56 &0.688&14   &0.544&43       \\

31& Synta  & Phrasal          &as\_PrPhr\_C& \# Prepositional Phrases per Sent  &0.513&35 &0.417&23 &0.607&70 &0.608&28   &0.590&34      \\

31& Synta  & Part-of-Speech   &as\_FuncW\_C& \# Function Words per Sent      &0.468&48 &0.471&8  &0.662&57 &0.673&17   &0.614&19       \\

31& LxSem  & Psycholinguistic &to\_AAKuL\_C& Total Sum Kuperman Lemma AoA       &0.428&71 &0.189&59 &0.835&3  &0.627&22   &0.716&5       \\

31& LxSem  & Psycholinguistic &to\_AAKuW\_C& Total Sum Kuperman Word AoA       &0.427&72 &0.189&60 &0.835&4  &0.625&23   &0.715&6       \\

30& LxSem  & Type Token Ratio & CorrTTR\_S & Corrected Type Token Ratio     &0.745&1  &0.006&228 &0.846&1  &0.445&65   &0.692&7       \\

\textit{30}& \textit{LxSem}  & \textit{Variation Ratio} & \textit{CorrNoV\_S} & \textit{Corrected Noun Variation-1}     &0.717&3  &0.0858&131 &0.842&2  &0.406&78   &0.612&21       \\

30& Synta  & Tree Structure   &as\_TreeH\_C& Tree Height per Sent               &0.550&21 &0.341&30 &0.686&51 &0.699&9    &0.541&44      \\

30& Synta  & Phrasal          &to\_PrPhr\_C& Total \# Prepositional Phrases   &0.470&47 &0.189&58 &0.808&11 &0.580&36   &0.729&3       \\

30& LxSem  & Word Familiarity &as\_SbL1C\_C& SubtlexUS Lg10CD Value per Sent  &0.467&49 &0.430&20 &0.612&69 &0.699&10   &0.533&45       \\

\cmidrule(lr){1-15}
\end{tabular}
}
\caption{\label{Table 4} Top 15 (score $\geq$ 30) handcrafted linguistic features under Approach B. Italic for the feature not in Table 3.}
\vspace{-4mm}
\end{table*}

\subsection{Feature Extraction \& Ranking}
We utilize LingFeat for feature extraction. It is a public software that supports 255 handcrafted linguistic features in the branches of advanced semantic, discourse, syntactic, lexico-semantic, and shallow traditional. They further classify into 14 subgroups. We study the linguistically-meaningful branches: discourse (entity density, entity grid), syntax (phrasal, tree structure, part-of-speech), and lexico-semantics (variation ratio, type token ratio, psycholinguistics, word familiarity). 

After extracting the features from CCB, WBT, CAM, CKC, and OSE, we first create feature performance ranking by Pearson's correlation. We used Sci-Kit Learn \citep{scikit-learn}. We take extra measures (Approach A \& B) to model the features' general performances across datasets. Each approach runs under differing premises:

\noindent
\textbf{Premise A}: "Human experts' dataset creation and labeling are partially faulty. The weak performance of a feature in a dataset does not necessarily indicate its weak performance in other data settings".

\noindent
\textbf{Premise B}: "All datasets are perfect. The weak performance of a feature in a dataset indicates the feature's weakness to be used universally." 

After 78 hours of running, we decided not to extract features from NSL. Computing details are in appendix E. Among the features included in LingFeat, there are traditional readability formulas, like FKGL and COLE. These formulas performed generally well but a single killer feature, like type token ratio (TTR), often outperformed formulas. Traditional readability formulas and shallow traditional features are excluded from the rankings.

\subsection{Approach A - Comparative Ranking}
Under premise A, each dataset poses a different linguistic environment to feature performance. Further, premise A takes human error into consideration and agrees that data labeling is most likely inconsistent in some way. The literal correlation value itself is not too important under premise A. 

\begin{table*}
\begin{align*}
  \text{\textbf{NERF}} 
  &= \text{(analogous to) \textbf{Lexical Difficulty}} + \text{\textbf{Syntactic Complexity}} + \text{\textbf{Lexical Richness}} + \text{\textbf{Bias}}\\[10pt] 
  &= \frac{0.04876 \cdot \sum \text{Word Age-of-Acquisition} -0.1145 \cdot \sum \text{Word Familiarity}}{\text{\#Sentence}}\\[10pt] 
  &+ \frac{0.3091 \cdot \text{\#Content Word} + 0.1866 \cdot \text{\#Noun Phrase} + 0.2645 \cdot \text{Constituency Parse Tree Height}}{\text{\#Sentence}}\\[10pt] 
  &+ \frac{1.1017 \cdot \text{\#Unique Word}}{\sqrt{\text{\#Word}}} - 4.125
\end{align*}
\caption*{Equation) New English Readability Formula (NERF)}
\vspace{-4mm}
\end{table*}

Rather, we look for features that perform better than the others, under the same test settings. Thus, approach A's rewarding system is rank-dependent. In a dataset, features that rank 1-10 are rewarded 10 points, rank 11-20 get 9 points, ... and rank 91-100 get 1 point. Since there are five feature correlation rankings (one per dataset), the maximum score is 50. The results are in Table 3, in the order of score.

\subsection{Approach B - Absolute Correlation}

Under premise B, the weak correlation of a feature in a dataset is solely due to the feature's weakness to generalize. This is because all datasets are supposedly perfect. Hence, we only measure the feature's absolute correlation across datasets. 

Approach B's rewarding system is correlation-dependent. In a dataset, features that show correlation value between 0.9-10 are rewarded 10 points, value between 0.8-0.89 get 9 points, ... and value between 0.0-0.09 get 1 point. Like approach A, the maximum score is 50. The result is in Table 4.

\subsection{Analysis \& Manual Feature Selection}
First and the most noticeable, the top features under premise A \& B are similar. In fact, the two results are almost replications of each other except for minor changes in order. We initially set two premises to introduce differing views (and hence the results) to feature rankings. Then, we would choose the features that perform well in both. 

But there seems to be an inseparable correlation between ranking-based (premise A) and correlation-based (premise B) approaches. CorrNoV\_S (Corrected Noun Variation) was the only new top feature introduced under premise B.

Second, discourse-based features (mostly entity-related) performed poorly for use in our final NERF. As an exception, ra\_NNToT\_C (noun-noun transitions : total) scored 28 under premise A and 26 under premise B. On the other hand, a majority of lexico-semantic and syntactic features performed well throughout. This strongly suggests that a possible discovery of universally-effective features for readability is in lexico-semantics or syntax.

Third, the difficulty of a document heavily depended on the difficulty of individual words. In detail, as\_AAKuL\_C, as\_AAKuW\_C, to\_AAKuL\_C, to\_AAKuW\_C showed consistently high correlations across the five datasets. As shown in Section 2, these five datasets have different authors, target audience, average length, labeling techniques, and the number of classes. Each dataset had at least one of these features among the top 5 performances.

The four features come from age-of-acquisition research by \citet{kuperman:2012}, which now prove to be an important resource for RA. Such direct classification of word difficulties always outperformed frequency-based approaches like SubtlexUS \citep{brysbaert:2009}. Back to feature selection, we follow the steps below.

\noindent
\textbf{1.} From top to bottom, go through ranking (table 3 \& 4) to sort out the features that performed the best in each linguistic subgroup.

\noindent
\textbf{2.} Conduct step 1 to both datasets and compare the results to each other. Though this process, we only leave the features that duplicate in both rankings.

The steps above produce the same results for both approach A and B. The final selected features are as\_AAKuL\_C (psycholinguistic), as\_TreeH\_C (tree structure), as\_ContW\_C (part-of-speech), as\_NoPhr\_C (phrasal), as\_SbL1C\_C (word familiarity), CorrTTR\_S (type token ratio). CorrNov\_S (variation) only appeared under approach B, and we did not include it.
\vspace{-1mm}
\subsection{More on NERF \& Calibration}
The final NERF (section 4.5) is brought in three parts. The first is lexico-semantics, which measures lexical difficulty. It adds the total sum of each word's age-of-acquisition (Kuperman's) and the sum of word familiarity scores (Lg10CD in SubtlexUS). The sum is divided by \# sentences.

The second is syntactic complexity, which deals with how each sentence is structured. We look at the number of content words, noun phrases, and the total sum of sentence tree height. Here, content words (CW) are words that possess semantic content and contribute to the meaning of the specific sentence. Following LingFeat, we consider a word to be a content word if it has "NOUN", "VERB", "NUM", "ADJ", "ADV" as a POS tag. Also, a sentence's tree height (TH) is calculated from a constituency-parsed tree, which we used the CRF parser \citep{zhang:2020} to obtain. The related algorithms from NLTK \citep{nltk} were used in calculating tree height. The same CRF parser was also used to count the number of noun phrase (NP) occurrences.

The third is lexical richness, given through type token ratio (TTR). This is the only section of NERF that is averaged on the word count. TTR measures how many unique vocabularies appear with respect to the total word count. TTR is often used as a measure of lexical richness \citep{malvern:2012} and ranked the best performance on two native datasets (CCB and CAM). Importantly, these two datasets represent US and UK school curriculums, and TTR seems a good evaluator. What was interesting is that out of the five TTR variations from \citet{lee:2021, weebit}, corrected TTR generalized particularly well.

Like section 3, we use the non-linear least fitting method on CCB to calibrate NERF. The results match what we expected. For example, the coefficient for word familiarity, which measures how frequently the word is used in American English, is negative since common words often have faster lexical comprehension times \citep{brysbaert:2011}.
\vspace{-1mm}
\section{Evaluation, against Human}
\vspace{-1mm}
\begin{table}[t]
\centering
\resizebox{0.5\textwidth}{!}{%
\begin{tabular}{
l@{\hspace{0.8ex}}
c@{\hspace{0.8ex}}
c@{\hspace{0.8ex}}
c@{\hspace{0.8ex}}
c@{\hspace{0.8ex}}
c@{\hspace{0.8ex}}
c@{\hspace{0.8ex}}
c}
\cmidrule(lr){1-8}
\textbf{Metric}&\textbf{Human}&\textbf{NERF}&\textbf{FKGL}&\textbf{FOGI}  &\textbf{SMOG} &\textbf{COLE} &\textbf{AUTO}\\ 
\cmidrule(lr){1-1}\cmidrule(lr){2-2}\cmidrule(lr){3-3}\cmidrule(lr){4-4}\cmidrule(lr){5-5}\cmidrule(lr){6-6}\cmidrule(lr){7-7}\cmidrule(lr){8-8}
MAE&N.A.&N.A.& 2.844& 3.413& 3.114& 2.537& 3.377  \\
\textbf{MAE}&3.509&2.154&2.457&2.516&2.728&2.378&2.514  \\
r2 score&N.A.&N.A.& -0.03835& -0.3905& 0.1613& 0.4341& -0.5283  \\
\textbf{r2 score}&-0.0312&0.5536&0.4423&0.4072&0.3192&0.4830&0.4263  \\
Pearson r&N.A.&N.A.& 0.5698& 0.5757& 0.5649& 0.6800& 0.5684  \\
\textbf{Pearson r}&0.0838&0.7440&0.6651&0.6381&0.5649&0.6949&0.6530  \\
\cmidrule(lr){1-8}
\end{tabular}
}
\caption{\label{Table 5} Scores on CCB. Measured on U.S. Standard Curriculum’s
K-* Output. Bold for new or adjusted.}
\vspace{-4mm}
\end{table}

Here, we check the human-perceived difficulty of each item in CCB. We used Amazon Mechanical Turk to ask U.S. Bachelor's degree holders, "Which U.S. grade does this text belong to?" Every item was answered by $10$ different workers to ensure breadth. Details on survey \& datasets are in appendix B, C. 

Table 5 gives a performance comparison of NERF against other traditional readability formulas and human performances. The human predictions were made by the U.S. Bachelor's degree holders living in the U.S. Ten human predictions were averaged to obtain the final prediction for each item, for comparison against CCB.

The calibrated formulas show a particularly great increase in r2 score. This likely means that the new recalibrated formulas can capture the variance of the original CCB classifications much better when compared to the original formulas. We believe that such an improvement stems from the change in datasets. The original formulas are mostly built on human tests of 20th century’s military or technical documents, whereas the recalibration dataset (CCB) are from the student-targeted school curriculum. Further, CCB is classified by trained professionals. Hence, the standards for readability can differ. The new recalibrated versions are more suitable for analyzing the modern general documents and giving K-* output by modernized standards.

MAE (Mean Absolute Error), r2 score, and Pearson's r improve once more with NERF. Even though the same dataset, same fitting function, and same evaluation techniques (no split, all train) were used, the critical difference was in the features. The shallow surface-level features from the traditional readability formulas also showed top rankings across all datasets but lacked linguistic coverage. Hence, NERF could capture more textual properties that led to a difference in readability.

Lastly, we observe that it is highly difficult for the general human population to exactly guess the readability of a text. Out of 690 predictions, only 286 were correct. We carefully posit that this is because: 1. the concept of "readability" is vague and 2. everyone goes through varying education. It could be easier to choose which item is more readable, instead of guessing how readable an item is. Given the general population, it is always better to use some quantified models than trust human.

\section{Evaluation, for Application}
\subsection{Text Simplification - Passage-based}
All readability formulas, whether recalibrated or not, show near-perfect performances in ranking the simplicity of texts. On both OSE-Pair \& NSL-Pair, we designed a simple task of ranking the simplicity of an item. Both paired datasets include multiple simplified versions of an original item. Each row consists of various simplifications. A correct prediction is the corresponding readability formula output matching simplification level (e.g. original: highest prediction, ..., simplest: lowest prediction).

In OSE-Pair, a correct prediction must properly rank three simplified items. NERF showed a meaningfully improved performance than the other five traditional readability formulas before recalibration. NERF correctly classified 98.7\% pairs, while the others stayed $\leq$95\% (FKGL: 93.4\%, FOGI: 92.6\%, SMOG: 94.4\%, COLE: 94.9\%, AUTO: 92.6\%). Recalibration generally helped the traditional readability formulas but NERF still showed better performance (FKGL: 97.8\%, FOGI: 97.1\%, SMOG: 94.4\%, COLE: 89.9\%, AUTO: 95.8\%).

In NSL-Pair, a correct prediction must properly rank five simplified items, which is a more difficult task than the previous. Nonetheless, all six formulas achieved 100\% accuracies. The same results were achieved before and after CCB-recalibration. This hints that NSL-Pair is thoroughly simplified.

Readability formulas seem to perform well in ranking several simplifications on a passage-level. But there certainly are limits. First, one must understand that calculating "how much simple" is a much difficult task (Table 5). Second, the good results could be because sufficient simplification was done. For more fine grained simplifications, readability formulas could not be enough.

\subsection{Text Simplification - Sentence-based}
\begin{table}[t]
\centering
\footnotesize
\begin{tabular}{
l@{\hspace{0.8ex}}
c@{\hspace{0.8ex}}
c@{\hspace{0.8ex}}
c@{\hspace{0.8ex}}
c@{\hspace{0.8ex}}
c@{\hspace{0.8ex}}
c}
\cmidrule(lr){1-7}
\textbf{a) Adv-Ele}&\textbf{NERF}&\textbf{FKGL}&\textbf{FOGI}  &\textbf{SMOG} &\textbf{COLE} &\textbf{AUTO}\\ 
\cmidrule(lr){1-1}\cmidrule(lr){2-2}\cmidrule(lr){3-3}\cmidrule(lr){4-4}\cmidrule(lr){5-5}\cmidrule(lr){6-6}\cmidrule(lr){7-7}
Accuracy         &N.A.  &74.2\%&64.9\%&11.4\%&66.0\%&78.0\%  \\
\textbf{Accuracy}&77.4\%&62.7\%&51.8\%&11.4\%&71.1\%&65.2\%  \\
\cmidrule(lr){1-7}

\textbf{b) Adv-Int}&\textbf{NERF}&\textbf{FKGL}&\textbf{FOGI}  &\textbf{SMOG} &\textbf{COLE} &\textbf{AUTO}\\ 
\cmidrule(lr){1-1}\cmidrule(lr){2-2}\cmidrule(lr){3-3}\cmidrule(lr){4-4}\cmidrule(lr){5-5}\cmidrule(lr){6-6}\cmidrule(lr){7-7}
Accuracy         &N.A.  &70.2\%&63.0\%&12.2\%&63.6\%&74.7\%  \\
\textbf{Accuracy}&77.8\%&60.4\%&51.3\%&12.2\%&67.7\%&65.9\%  \\
\cmidrule(lr){1-7}

\textbf{c) Int-Ele}&\textbf{NERF}&\textbf{FKGL}&\textbf{FOGI}  &\textbf{SMOG} &\textbf{COLE} &\textbf{AUTO}\\ 
\cmidrule(lr){1-1}\cmidrule(lr){2-2}\cmidrule(lr){3-3}\cmidrule(lr){4-4}\cmidrule(lr){5-5}\cmidrule(lr){6-6}\cmidrule(lr){7-7}
Accuracy         &N.A.  &69.8\%&61.3\%&9.02\%&61.9\%&73.2\%  \\
\textbf{Accuracy}&73.1\%&59.7\%&48.9\%&9.02\%&66.5\%&62.1\%  \\
\cmidrule(lr){1-7}

\end{tabular}

\caption{\label{Table 6} Scores on OSE-Sent. Bold for new or adjusted.}
\vspace{-4mm}
\end{table}
We were surprised that some existing text simplification studies are directly using traditional readability formulas for sentence difficulty evaluation. Our results show that using a formula-based approach is particularly useless in evaluating a sentence. 

We tested both CCB-recalibrated and original formulas on ASSET. Here, a correct prediction must properly rank eleven simplified items. Despite the task difficulty, we anticipated seeing some correct predictions as there were 360 pairs. SMOG guessed 37 (after recalibration) and 89 (before recalibration) correct out of 360. But all the other formulas failed to make any correct prediction. 

OSE-Sent poses an easier task. Since the dataset is divided into adv-int, adv-ele, and int-ele, the readability formulas now had to guess which is more difficult, out of the given two. We do obtain some positive results, showing that readability formulas can be useful in the cases where only two sentences are compared. On ranking two sentences, NERF performs better by a large margin.

\subsection{Medical Documents}
\begin{figure}
    \includegraphics[width=0.5\textwidth]{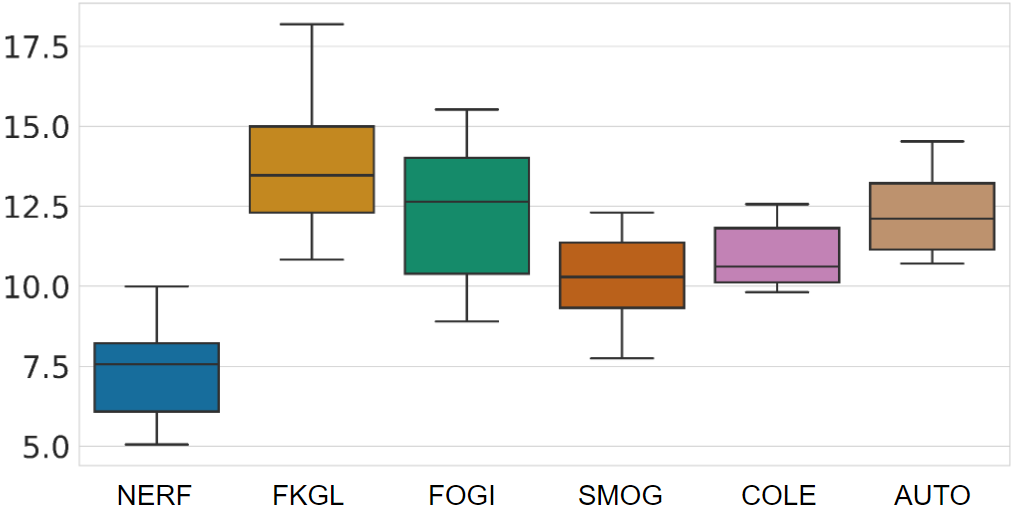}
    \caption{On medical texts. NERF, against five others.}
    \vspace{-4mm}
\end{figure}
We argue that NERF is effective in fixing the over-inflated prediction of difficulty on medical texts. Such sudden inflation is widely-reported \citep{zheng2017readability} as the common weaknesses of traditional readability formulas on medical documents.

The U.S. National Institute of Health (NIH) guides that patient documents be $\leq$K-6 of difficulty. The most distinct characteristic of medical documents is the use of lengthy medical terms, like otolaryngology, urogynecology, and rheumatology. This makes traditional formulas, based on syllables, unreliable. But NERF uses familiarity and age-of-acquisition to penalty and reward word difficulty.

A medical term not found in Kuperman's and SubtlexUS will have no effect. Instead, it will simply be labeled a content word. But in traditional formulas, the repetitive use of medical terms (which is likely the case) results in an insensible aggregation of text difficulty. In case various medical terms appear, NERF rewards each as a unique word.

Among recent studies is \citet{haller:2019}, which analyzed the readability of urogynecology patient education documents in FKGL, SMOG, and Fry Readability. We also analyze the same 18 documents from the American Urogynecologic Society (AUGS) by manual OCR-based scraping. As Figure 1 shows, it is evident that NERF helps regulate the traditional readability formulas' tendencies to over-inflate on medical texts. An example of the collected resource is given in appendix B.

\section{Conclusion}
So far, we have recalibrated five traditional readability formulas and assessed their performances. We evaluated them on CCB and proved that the adjusted variations help traditional readability formulas give output more in align with CCB, a common English education curriculum used throughout the United States. Further, we evaluated the recalibrated formulas' application on text simplification research. On ranking passage difficulty, our recalibrated formulas showed good performance. However, the formulas lacked performance on ranking sentence difficulty because they were calibrated on passage-length instances. We leave sentence difficulty ranking as an open task.

Apart from recalibration traditional readability formulas, we also develop a new, linguistically-rich readability formulas named NERF. We prove that NERF can be much more useful when it comes to text simplification studies and analyzing the readability of medical documents. Also, our paper serves as a cross-comparison among readability metrics. Lastly, we develop a public Python-based software, for the fast dissemination of the results.

\section{Limitations}
Our work's limitations mainly come from CCB. It is manifestly difficult to obtain solid, gold readability-labelled dataset from an officially accredited organization. CCB, the main dataset that we used to calibrate traditional readability formulas, has only 69 items available. Thus, we reasonably anticipate that variation in dialect, individual differences and general ability cannot be captured. 

However, we highlight that NERF is developed upon several more datasets that represent diverse background, audience, and reading level. Hence, we believe that NERF can counter some of the shallowness of the traditional readability formulas, despite the still existing weaknesses.

One aspect of readability formulas that have not been deeply investigated is how the output changes depending on the text length. As we show in section 7, readability formulas fail to perform well on sentence-level items. But how about a passage of three sentences? Or does the performance have to do with the average number of words in the recalibration dataset? Is there some sensible range that the readability formulas work well for? These are some open question we fail to address in this work.

\clearpage
\bibliography{anthology,custom}
\bibliographystyle{acl_natbib}
\vspace{10mm}
\appendix
\section{Public Resources We Developed}
\subsection{Python Library}
\subsubsection{As a Readability Tool}
<Anonymous> supports six readability formulas: NERF, FKGL, FOGI, SMOG, COLE and AUTO. All formulas, other than NERF, are also available in recalibrated variations. A particularly useful feature of this library is that all formulas are fitted to give the U.S. standard school grading system as output. Compared to some other traditional readability formulas where a user has to refer to a table understand output, K-* based numbers are intuitive.

\subsubsection{As a General Tool}
We have plans to expand <Anonymous> to support various menial tasks in text analysis. We are to focus on the tasks that can be better performed using simplistic approaches. One feature that we had already implemented is text reading time estimation. \citet{weller:2020} has previously shown in a large-scale study that a commonly used rule-of-thumb for online reading estimates, 240 words per minute (WPM), shows better RMSE and MAE results when compared to more modern approaches using XLNet \citep{yang2019xlnet}, ELMo \citep{peters-etal-2018-deep} and RoBERTa \citep{liu2019roberta}. We implement 175, 240 and 300 WPM. 

\subsubsection{Basic Usage}
For straightforward maintenance, we keep <Anonymous>'s architecture as simple as possible. There are not many steps for the user to take:

\noindent
\textit{import <Anonymous>}

\noindent
\textit{new\_object = <Anonymous>.request(...)}

\noindent
\textit{readability\_score1 = new\_object.NERF()}

\noindent
\textit{readability\_score2 = new\_object.FKGL()}

\noindent
\textit{readability\_score3 = new\_object.FOGI()}

\noindent
\textit{readability\_score4 = new\_object.SMOG()}

\noindent
\textit{readability\_score5 = new\_object.COLE()}

\noindent
\textit{readability\_score6 = new\_object.AUTO()}

\noindent
\textit{time\_to\_read = new\_object.RT()}

NERF(), FKGL(), FOGI(), SMOG(), COLE(), AUTO(), RT() are shortcut functions. It can be \textit{slightly} faster to directly call in the full forms as:

\noindent
\textit{new\_english\_readability\_formula() }

\noindent
\textit{flesch\_kincaid\_grade\_level()}

\noindent
\textit{fog\_index()}

\noindent
\textit{smog\_index()}

\noindent
\textit{coleman\_liau\_index()}

\noindent
\textit{automated\_readability\_index()}

\noindent
\textit{read\_time()}

Further, all readability formula functions (except for NERF) has option to choose the original or the adjusted variation. Default is set \textit{adjusted = True}. 

\subsubsection{<Anonymous> Speed to Calculation}
We care for the library's calculation speed so that it can be of practical use for research implementations. We chose the following items for evaluation.

\noindent
\textbf{ITEM A}

\noindent
In those times panics were common, and few days passed without some city or other registering in its archives an event of this kind. There were nobles, who made war against each other; there was the king, who made war against the cardinal; there was Spain, which made war against the king. Then, in addition to these concealed or public, secret or open wars, there were robbers, mendicants, Huguenots, wolves, and scoundrels, who made war upon everybody. The citizens always took up arms readily against thieves, wolves or scoundrels, often against nobles or Huguenots, sometimes against the king, but never against the cardinal or Spain. It resulted, then, from this habit that on the said first Monday of April, 1625, the citizens, on hearing the clamor, and seeing neither the red-and-yellow standard nor the livery of the Duc de Richelieu, rushed toward the hostel of the Jolly Miller. When arrived there, the cause of the hubbub was apparent to all.

\noindent
\textit{The Three Musketeers, Alexandre Dumas}

\noindent
\textbf{ITEM B}

\noindent
The vaccine contains lipids (fats), salts, sugars and buffers. COVID-19 vaccines do not contain eggs, gelatin (pork), gluten, latex, preservatives, antibiotics, adjuvants or aluminum. The vaccines are safe, even if you have food, drug, or environmental allergies. Talk to a health care provider first before getting a vaccine if you have allergies to the following vaccine ingredients: polyethylene glycol (PEG), polysorbate 80 and/or tromethamine (trometamol or Tris).

\noindent
\textit{COVID-19 Vaccine Information Sheet, Ministry of Health, Ontario Canada}

\noindent
\textbf{ITEM C}

\noindent
BERT alleviates the previously mentioned unidirectionality constraint by using a “masked language model”(MLM) pre-training objective, inspired by the Cloze task.

\noindent
\textit{Pre-training of Deep Bidirectional Transformers for Language Understanding, Jacob Devlin, Ming-Wei Chang, Kenton Lee, Kristina Toutanova}

\setcounter{table}{7}
\begin{table}[!hbtp]
\centering
\resizebox{0.48\textwidth}{!}{%
\begin{tabular}{
l@{\hspace{0.8ex}}
c@{\hspace{0.8ex}}
c@{\hspace{0.8ex}}
c@{\hspace{0.8ex}}
c@{\hspace{0.8ex}}
c@{\hspace{0.8ex}}
c}
\cmidrule(lr){1-7}
\textbf{a) ITEM A}&\textbf{NERF}&\textbf{FKGL}&\textbf{FOGI}  &\textbf{SMOG} &\textbf{COLE} &\textbf{AUTO}\\ 
\cmidrule(lr){1-1}\cmidrule(lr){2-2}\cmidrule(lr){3-3}\cmidrule(lr){4-4}\cmidrule(lr){5-5}\cmidrule(lr){6-6}\cmidrule(lr){7-7}
item * 1         &0.6371&0.0002&0.0001&0.0001&0.0000&0.0000  \\
item * 5         &2.6450&0.0006&0.0005&0.0004&0.0001&0.0001  \\
item * 10        &5.5175&0.0011&0.0010&0.0010&0.0004&0.0004  \\
item * 15        &7.8088&0.0016&0.0016&0.0013&0.0003&0.0004  \\
item * 20        &10.226&0.0021&0.0021&0.0018&0.0004&0.0004  \\
\cmidrule(lr){1-7}

\textbf{b) ITEM B}&\textbf{NERF}&\textbf{FKGL}&\textbf{FOGI}  &\textbf{SMOG} &\textbf{COLE} &\textbf{AUTO}\\ 
\cmidrule(lr){1-1}\cmidrule(lr){2-2}\cmidrule(lr){3-3}\cmidrule(lr){4-4}\cmidrule(lr){5-5}\cmidrule(lr){6-6}\cmidrule(lr){7-7}
item * 1         &0.3531&0.0000&0.0000&0.0000&0.0000&0.0000  \\
item * 5         &1.2842&0.0003&0.0003&0.0002&0.0000&0.0000  \\
item * 10        &2.5178&0.0005&0.0005&0.0004&0.0001&0.0001  \\
item * 15        &3.6545&0.0009&0.0007&0.0006&0.0002&0.0002  \\
item * 20        &4.8308&0.0010&0.0010&0.0009&0.0002&0.0002  \\
\cmidrule(lr){1-7}

\textbf{c) ITEM C}&\textbf{NERF}&\textbf{FKGL}&\textbf{FOGI}  &\textbf{SMOG} &\textbf{COLE} &\textbf{AUTO}\\ 
\cmidrule(lr){1-1}\cmidrule(lr){2-2}\cmidrule(lr){3-3}\cmidrule(lr){4-4}\cmidrule(lr){5-5}\cmidrule(lr){6-6}\cmidrule(lr){7-7}
item * 1         &0.1373&0.0000&0.0000&0.0000&0.0000&0.0000  \\
item * 5         &0.1888&0.0001&0.0000&0.0000&0.0000&0.0000  \\
item * 10        &0.2528&0.0002&0.0002&0.0002&0.0000&0.0000  \\
item * 15        &0.3420&0.0003&0.0003&0.0002&0.0000&0.0000  \\
item * 20        &0.3886&0.0004&0.0003&0.0003&0.0000&0.0000  \\
\cmidrule(lr){1-7}

\end{tabular}
}
\caption{\label{Table 7} Speeds in seconds, on Items A, B and C.}
\vspace{-4mm}
\end{table}

First, it is very obvious that AUTO does a great job in keeping calculation speed short for longer texts as originally intended. Second, NERF's calculation speed linearly increases in respect to the text length. Though, we believe that NERF's speed is decent in its wide linguistic coverage, it seems true that the speed is weakness when compared to the other readability formulas.

\subsection{Research Archive}
Our datasets, preprocessing codes and evaluation codes can be found in <Anonymous>. Copyrighted resources are given upon request to the first author.

\section{External Resources}

\subsection{Python Libraries}
\noindent
\textbf{pandas} v.1.3.4 \citep{pandas}

\noindent
Calculations for Kuperman's AoA CSV, SubtlexUS word familiarity CSV, manage and manipulate data. For feature study purposes, correlate and rank features in Tables 3 and 4.

\noindent
\textbf{SuPar} v.1.1.3 - CRF Parser

\noindent
Constituency parsing on input sentences -> calculate tree height and count noun phrases.

\noindent
\textbf{spaCy} v.3.2.0 \citep{spacy}

\noindent
Sentence/dependency parsing on documents -> sent input into SuPar and count content words (POS).

\noindent
\textbf{Sci-Kit Learn} v.1.0.1

\noindent
Calculation, r2 score and MAE in Tables 2 and 5.

\noindent
\textbf{SciPy} v.1.7.3

\noindent
Calculation of Pearson's r for Tables 2 and 5. Fitting function (scipy.optimize.curve\_fit()) used to recalibrate traditional readability formulas and give coefficients for NERF in Table 2.

\noindent
\textbf{NLTK} v.3.6.5

\noindent
Calculation of tree height for NERF.

\noindent
\textbf{LingFeat} v.1.0.0-beta.19

\noindent
Extraction of handcrafted linguistic features.

\subsection{Datasets}

\begin{table}[!htbp]
\centering
\resizebox{0.48\textwidth}{!}{%
\begin{tabular}{l|ll}
\hline\textbf{New Class}    & \textbf{CCB} & \textbf{WBT} \\ 
\cmidrule(lr){1-1}\cmidrule(lr){2-2}\cmidrule(lr){3-3}
K1.0                  & K1 (Age 6-7)          & N/A            \\     
K2.0                  & N/A                   & Level 2 (Age 7-8)      \\
K2.5                  & K2-3 (Age 7-9)        & N/A          \\
K3.0                  & N/A                   & Level 3 (Age 8-9)      \\
K4.0                  & N/A                   & Level 4 (Age 9-10)    \\
K4.5                  & K4-5 (Age 9-11)       & N/A         \\
K7.0                  & K6-8 (Age 11-14)      & KS3 (Age 11-14)   \\
K9.5                 & K9-10 (Age 14-16)     & GCSE (Age 14-16)\\
K12.0                & K11-CCR (Age 16+)     & N/A\\
\cmidrule(lr){1-3}
\end{tabular}
}
\caption{\label{Table 8} Aged-based conversions for CCB and WBT.}
\vspace{-4mm}
\end{table}

We collected CCB by manually going through an official source\footnote{corestandards.org/assets/Appendix\_B.pdf}. WBT was obtained from the authors\footnote{Dr. Sowmya Vajjala, National Research Council, Canada} in HTML format. We conducted basic preprocessing and manipulated WBT in CSV format. CAM was retrieved from an existing archive\footnote{ilexir.co.uk/datasets/index.html}. CKC was retrieved from a South Korean educational company\footnote{Bruce W. Lee, LXPER Inc., South Korea}. OSE was retrieved from a public archive\footnote{github.com/nishkalavallabhi/OneStopEnglishCorpus}. NSL was obtained from an American educational company\footnote{Luke Orland, Newsela Inc., New York, U.S.A.}. AUGS medical texts (refer to Section 6.3) were manually scraped from the official website\footnote{augs.org/patient-fact-sheets/}. ASSET was obtained from a public repository\footnote{github.com/facebookresearch/asset}. Lastly, Table 8 shows how we converted WBT class labels to fit CCB and show in Table 1. All were consistent with intended use.

Further, to give more backgrounds to section 6.2, we give example pairs from ASSET and OSE-Sent.

\noindent
\textbf{ASSET}

\noindent
\textbf{0}: Gable earned an Academy Award nomination for portraying Fletcher Christian in Mutiny on the Bounty.

\noindent
\textbf{1}: Gable also earned an Oscar nomination when he portrayed Fletcher Christian in 1935's Mutiny on the Bounty.

\noindent
\textbf{2}: Gable won an Academy Award vote when he acted in 1935's Mutiny on the Bounty as Fletcher Christian.

\noindent
\textbf{3}: Gable also won an Academy Award nomination when he played Fletcher Christian in the 1935 film Mutiny on the Bounty.

\noindent
\textbf{4}: Gable was nominated for an Academy Award for portraying Fletcher Christian in 1935's Mutiny on the Bounty.

\noindent
\textbf{5}: Gable also earned an Academy Award nomination in 1935 for playing Fletcher Christian in "Mutiny on the Bounty.

\noindent
\textbf{6}: Gable also earned an Academy Award nomination when he played Fletcher Christian in 1935's Mutiny on the Bounty.

\noindent
\textbf{7}: Gable recieved an Academy Award nomination for his role as Fletcher Christian. The film was Mutiny on the Bounty (1935).

\noindent
\textbf{8}: Gable earned an Academy Award nomination for his role as Fletcher Christian in the 1935 film Mutiny on the Bounty.

\noindent
\textbf{9}: Gable also got an Academy Award nomination when he played Fletcher Christian in 1935's movie, Mutiny on the Bounty.

\noindent
\textbf{10}: Gable also earned an Academy Award nomination when he portrayed Fletcher Christian in 1935's Mutiny on the Bounty.\\

\noindent
\textbf{OSE-Sent (ADV-ELE)}

\noindent
\textbf{ADV}: The Seattle-based company has applied for its brand to be a top-level domain name (currently .com), but the South American governments argue this would prevent the use of this internet address for environmental protection, the promotion of indigenous rights and other public interest uses. 

\noindent
\textbf{ELE}: Amazon has asked for its company name to be a top-level domain name (currently .com), but the South American governments say this would stop the use of this internet address for environmental protection, indigenous rights and other public interest uses.\\

\noindent
\textbf{OSE-Sent (ADV-INT)}

\noindent
\textbf{ADV}: Brazils latest funk sensation, Anitta, has won millions of fans by taking the favela sound into the mainstream, but she is at the centre of a debate about skin colour.

\noindent
\textbf{INT}: Brazils latest funk sensation, Anitta, has won millions of fans by making the favela sound popular, but she is at the centre of a debate about skin colour.\\

\noindent
\textbf{OSE-Sent (INT-ELE)}

\noindent
\textbf{INT}: Allowing private companies to register geographical names as gTLDs to strengthen their brand or to profit from the meaning of these names is not, in our view, in the public interest, the Brazilian Ministry of Science and Technology said.

\noindent
\textbf{ELE}: Allowing private companies to register geographical names as gTLDs to profit from the meaning of these names is not, in our view, in the public interest, the Brazilian Ministry of Science and Technology said.\\

The following is an example of the AUGS medical documents used in Section 6.3 and Figure 1.

\noindent
\textbf{Interstitial Cystitis}: Interstitial Cystitis/ Bladder Pain Syndrome Interstitial cystitis/bladder pain syndrome (IC/BPS) is a condition with symptoms including burning, pressure, and pain in the bladder along with urgency and frequency. About IC/BPS IC/BPS occurs in three to seven percent of women, and can affect men as well. Though usually diagnosed among women in their 40s, younger and older women have IC/BPS, too. It can feel like a constant bladder infection. Symptoms may become severe (called a "flare") for hours, days or weeks, and then disappear. Or, they may linger at a very low level during other times. Individuals with IC/BPS may also have other health issues such as irritable bowel syndrome, fibromyalgia, chronic headaches, and vulvodynia. Depression and anxiety are also common among women with this condition. The cause of IC/BPS is unknown. It is likely due to a combination of factors. IC/BPS runs in families and so may have a genetic factor. On cystoscopy, the doctor may see damage to the wall of the bladder. This may allow toxins from the urine to seep into the delicate layers of the bladder lining, causing the pain of IC/BPS. Other research found that nerves in and around the bladder of people with IC/BPS are hypersensitive. This may also contribute to IC/BPS pain. There may also be an allergic component. 

\section{CCB Human Predictions}
In Section 2.1, we mention that human predictions were collected on Amazon Mechanical Turk. Then, we compared human performance to readability formulas in Table 5. Here, surveys are designed.

\noindent
\textbf{Description}: must choose which difficulty level does the text belong, "difficulty does not correlate with text length"

\noindent
\textbf{Qualification Requirement(s)}: Location is one of US, HIT Approval Rate (\%) for all Requesters' HITs greater than 80, Number of HITs Approved greater than 50, US Bachelor's Degree equal to true, Masters has been granted

All 69 story-type items from CCB were given. Each item had to be completed by at least 10 different individuals, resulting in 690 responses in total. They were given 6 representative examples. Payments were adequately and they were informed that the responds shall be used for research.

\section{Handcrafted Linguistic Features and the Respective Generalizability}
We give full generalizability rankings that we obtained through LingFeat. Considering that much work has to be done on the generalizability of RA, we believe that these rankings are particularly helpful. Table 9, Table 10, Table 11, Table 12, Table 13, Table 14,Table 15 are expanded versions of Table 3 and Table 4. The features not shown scored a 0.

From the full rankings, it is clear that shallow traditional (surface-level), lexico-semantic and syntactic features are effective throughout all datasets. Advanced semantics and discourse features show some what similar mid-low performances. However, it should be acknowledged that among the worst performing are lexico-semantic and syntactic features, too. This is perhaps because LingFeat itself has a very lexico-semantics and syntax-focused collection of handcrafted linguistic features. Thus, more study is needed.

Even if two features are from the same group (phrasal), they could show drastically varying performances (\# Noun phrases per Sent - scored 39 in approach A v.s. \# Verb phrases per Sent - scored 1 in approach A). Hence, thorough feature study must always be conducted during research. In a feature selection for a readability-related model, a cherry picking the most well performing feature from each feature group is recommended.

\section{Computing Power}
Single CPU chip. \textbf{Architecture}: x86\_64; \textbf{CPU(s)}: 16; \textbf{Model name}: Intel(R) Core(TM) i9-9900KF CPU @ 3.60GHz; \textbf{CPU MHz}: 800.024

\begin{table*}[t]
\centering
\resizebox{\textwidth}{!}{%
\begin{tabular}{
c@{\hspace{0.8ex}}
l@{\hspace{0.8ex}}
l@{\hspace{0.8ex}}
l@{\hspace{0.8ex}}
l@{\hspace{0.8ex}}
|c@{\hspace{0.8ex}}
c@{\hspace{0.8ex}}
|c@{\hspace{0.8ex}}
c@{\hspace{0.8ex}}
|c@{\hspace{0.8ex}}
c@{\hspace{0.8ex}}
|c@{\hspace{0.8ex}}
c@{\hspace{0.8ex}}
|c@{\hspace{0.8ex}}
c@{\hspace{0.8ex}}}
\cmidrule(lr){1-15}
\multicolumn{5}{c}{\textbf{Feature}} & \multicolumn{2}{c}{\textbf{CCB}} & \multicolumn{2}{c}{\textbf{WBT}} & \multicolumn{2}{c}{\textbf{CAM}} & \multicolumn{2}{c}{\textbf{CKC}} & \multicolumn{2}{c}{\textbf{OSE}} \\
\cmidrule(lr){1-5}\cmidrule(lr){6-7}\cmidrule(lr){8-9}\cmidrule(lr){10-11}\cmidrule(lr){12-13}\cmidrule(lr){14-15}
Score & Branch & Subgroup & LingFeat Code & Brief Explanation          &   r  &  rk &  r &  rk &  r  &  rk & r  &  rk& r  &   rk  \\
\cmidrule(lr){1-1}\cmidrule(lr){2-2}\cmidrule(lr){3-3}\cmidrule(lr){4-4}\cmidrule(lr){5-5}\cmidrule(lr){6-6}\cmidrule(lr){7-7}\cmidrule(lr){8-8}\cmidrule(lr){9-9}\cmidrule(lr){10-10}\cmidrule(lr){11-11}\cmidrule(lr){12-12}\cmidrule(lr){13-13}\cmidrule(lr){14-14}\cmidrule(lr){15-15}
43&ShaTr&Shallow&as\_Sylla\_C&\# syllables per Sent &0.541&24   &0.461&10   &0.686&50   &0.697&11   &0.59&31\\
43&LxSem&Psycholinguistic&as\_AAKuL\_C&lemmas AoA of lemmas per Sent &0.54&25   &0.505&1   &0.722&42   &0.711&4   &0.601&25\\
43&ShaTr&Shallow&as\_Chara\_C&\# characters per Sent &0.539&27   &0.487&4   &0.696&46   &0.711&5   &0.613&20\\
43&LxSem&Psycholinguistic&as\_AAKuW\_C&AoA of words per Sent &0.537&28   &0.502&2   &0.722&41   &0.711&6   &0.602&24\\
42&Synta&Tree Structure&as\_FTree\_C&length of flattened Trees per Sent &0.505&37   &0.485&5   &0.677&54   &0.719&2   &0.622&16\\
40&LxSem&Psycholinguistic&at\_AAKuW\_C&AoA of words per Word &0.703&5   &0.308&36   &0.784&20   &0.643&21   &0.455&66\\
\end{tabular}
}
\caption{\label{Table 8} Part A. The full generalizability ranking of handcrafted linguistic features under Approach A. r: Pearson's correlation between the feature and the dataset. rk: the feature's correlation ranking on the specific dataset.}
\end{table*}

\begin{table*}[t]
\centering
\resizebox{\textwidth}{!}{%
\begin{tabular}{
c@{\hspace{0.8ex}}
l@{\hspace{0.8ex}}
l@{\hspace{0.8ex}}
l@{\hspace{0.8ex}}
l@{\hspace{0.8ex}}
|c@{\hspace{0.8ex}}
c@{\hspace{0.8ex}}
|c@{\hspace{0.8ex}}
c@{\hspace{0.8ex}}
|c@{\hspace{0.8ex}}
c@{\hspace{0.8ex}}
|c@{\hspace{0.8ex}}
c@{\hspace{0.8ex}}
|c@{\hspace{0.8ex}}
c@{\hspace{0.8ex}}}
\cmidrule(lr){1-15}
\multicolumn{5}{c}{\textbf{Feature}} & \multicolumn{2}{c}{\textbf{CCB}} & \multicolumn{2}{c}{\textbf{WBT}} & \multicolumn{2}{c}{\textbf{CAM}} & \multicolumn{2}{c}{\textbf{CKC}} & \multicolumn{2}{c}{\textbf{OSE}} \\
\cmidrule(lr){1-5}\cmidrule(lr){6-7}\cmidrule(lr){8-9}\cmidrule(lr){10-11}\cmidrule(lr){12-13}\cmidrule(lr){14-15}
Score & Branch & Subgroup & LingFeat Code & Brief Explanation          &   r  &  rk &  r &  rk &  r  &  rk & r  &  rk& r  &   rk  \\
\cmidrule(lr){1-1}\cmidrule(lr){2-2}\cmidrule(lr){3-3}\cmidrule(lr){4-4}\cmidrule(lr){5-5}\cmidrule(lr){6-6}\cmidrule(lr){7-7}\cmidrule(lr){8-8}\cmidrule(lr){9-9}\cmidrule(lr){10-10}\cmidrule(lr){11-11}\cmidrule(lr){12-12}\cmidrule(lr){13-13}\cmidrule(lr){14-14}\cmidrule(lr){15-15}
17&AdSem&Wiki Knowledge&WTopc05\_S&Number of topics, 50 topics extracted from Wiki &0.549&22   &0.033&186   &0.514&89   &0.533&50   &0.042&203\\
17&Synta&Part-of-Speech&as\_AvTag\_C&\# Adverb POS tags per Sent &0.32&129   &0.292&41   &0.526&83   &0.43&67   &0.415&79\\
16&LxSem&Type Token Ratio&BiLoTTR\_S&Bi-Logarithmic TTR &0.591&13   &0.062&149   &0.07&200   &0.001&229   &0.523&47\\
16&AdSem&Wiki Knowledge&WRich15\_S&Semantic Richness, 150 topics extracted from Wiki &0.495&39   &0.02&208   &0.48&95   &0.549&45   &0.037&209\\
16&Synta&Part-of-Speech&as\_CoTag\_C&\# Coordinating Conjunction POS tags per Sent &0.38&91   &0.411&24   &0.463&97   &0.442&66   &0.293&107\\
15&Synta&Phrasal&to\_AvPhr\_C&total \# Adverb phrases &0.356&105   &0.17&66   &0.705&45   &0.298&111   &0.432&73\\
\end{tabular}
}
\caption{\label{Table 9} Part B. The full generalizability ranking of handcrafted linguistic features under Approach A.}
\end{table*}

\begin{table*}[t]
\centering
\resizebox{\textwidth}{!}{%
\begin{tabular}{
c@{\hspace{0.8ex}}
l@{\hspace{0.8ex}}
l@{\hspace{0.8ex}}
l@{\hspace{0.8ex}}
l@{\hspace{0.8ex}}
|c@{\hspace{0.8ex}}
c@{\hspace{0.8ex}}
|c@{\hspace{0.8ex}}
c@{\hspace{0.8ex}}
|c@{\hspace{0.8ex}}
c@{\hspace{0.8ex}}
|c@{\hspace{0.8ex}}
c@{\hspace{0.8ex}}
|c@{\hspace{0.8ex}}
c@{\hspace{0.8ex}}}
\cmidrule(lr){1-15}
\multicolumn{5}{c}{\textbf{Feature}} & \multicolumn{2}{c}{\textbf{CCB}} & \multicolumn{2}{c}{\textbf{WBT}} & \multicolumn{2}{c}{\textbf{CAM}} & \multicolumn{2}{c}{\textbf{CKC}} & \multicolumn{2}{c}{\textbf{OSE}} \\
\cmidrule(lr){1-5}\cmidrule(lr){6-7}\cmidrule(lr){8-9}\cmidrule(lr){10-11}\cmidrule(lr){12-13}\cmidrule(lr){14-15}
Score & Branch & Subgroup & LingFeat Code & Brief Explanation          &   r  &  rk &  r &  rk &  r  &  rk & r  &  rk& r  &   rk  \\
\cmidrule(lr){1-1}\cmidrule(lr){2-2}\cmidrule(lr){3-3}\cmidrule(lr){4-4}\cmidrule(lr){5-5}\cmidrule(lr){6-6}\cmidrule(lr){7-7}\cmidrule(lr){8-8}\cmidrule(lr){9-9}\cmidrule(lr){10-10}\cmidrule(lr){11-11}\cmidrule(lr){12-12}\cmidrule(lr){13-13}\cmidrule(lr){14-14}\cmidrule(lr){15-15}
1&Synta&Part-of-Speech&ra\_SuVeT\_C&ratio of Subordinating Conjunction POS \# to Verb POS \# &0.36&103   &0.053&159   &0.109&177   &0.282&115   &0.137&139\\
1&Synta&Part-of-Speech&ra\_AjAvT\_C&ratio of Adjective POS \# to Adverb POS \# &0.357&104   &0.042&172   &0.056&204   &0.091&180   &0.044&199\\
1&LxSem&Psycholinguistic&at\_AABrL\_C&lemmas AoA of lemmas, Bristol norm per Word &0.333&124   &0.029&194   &0.462&98   &0.284&113   &0.217&118\\
1&LxSem&Psycholinguistic&at\_AACoL\_C&AoA of lemmas, Cortese and Khanna norm per Word &0.333&125   &0.029&193   &0.462&99   &0.284&114   &0.217&119\\
1&AdSem&WB Knowledge&BNois10\_S&Semantic Noise, 100 topics extracted from WeeBit  &0.193&171   &0.036&180   &0.37&110   &0.161&149   &0.33&97\\
1&AdSem&WB Knowledge&BNois05\_S&Semantic Noise, 50 topics extracted from WeeBit  &0.158&184   &0.011&219   &0.351&116   &0.15&153   &0.325&98\\
\end{tabular}
}
\caption{\label{Table 10} Part C. The full generalizability ranking of handcrafted linguistic features under Approach A.}
\end{table*}

\begin{table*}[t]
\centering
\resizebox{\textwidth}{!}{%
\begin{tabular}{
c@{\hspace{0.8ex}}
l@{\hspace{0.8ex}}
l@{\hspace{0.8ex}}
l@{\hspace{0.8ex}}
l@{\hspace{0.8ex}}
|c@{\hspace{0.8ex}}
c@{\hspace{0.8ex}}
|c@{\hspace{0.8ex}}
c@{\hspace{0.8ex}}
|c@{\hspace{0.8ex}}
c@{\hspace{0.8ex}}
|c@{\hspace{0.8ex}}
c@{\hspace{0.8ex}}
|c@{\hspace{0.8ex}}
c@{\hspace{0.8ex}}}
\cmidrule(lr){1-15}
\multicolumn{5}{c}{\textbf{Feature}} & \multicolumn{2}{c}{\textbf{CCB}} & \multicolumn{2}{c}{\textbf{WBT}} & \multicolumn{2}{c}{\textbf{CAM}} & \multicolumn{2}{c}{\textbf{CKC}} & \multicolumn{2}{c}{\textbf{OSE}} \\
\cmidrule(lr){1-5}\cmidrule(lr){6-7}\cmidrule(lr){8-9}\cmidrule(lr){10-11}\cmidrule(lr){12-13}\cmidrule(lr){14-15}
Score & Branch & Subgroup & LingFeat Code & Brief Explanation          &   r  &  rk &  r &  rk &  r  &  rk & r  &  rk& r  &   rk  \\
\cmidrule(lr){1-1}\cmidrule(lr){2-2}\cmidrule(lr){3-3}\cmidrule(lr){4-4}\cmidrule(lr){5-5}\cmidrule(lr){6-6}\cmidrule(lr){7-7}\cmidrule(lr){8-8}\cmidrule(lr){9-9}\cmidrule(lr){10-10}\cmidrule(lr){11-11}\cmidrule(lr){12-12}\cmidrule(lr){13-13}\cmidrule(lr){14-14}\cmidrule(lr){15-15}
35&LxSem&Psycholinguistic&as\_AAKuL\_C&lemmas AoA of lemmas per Sent &0.54&25   &0.505&1   &0.722&42   &0.711&4   &0.601&25\\
35&LxSem&Psycholinguistic&as\_AAKuW\_C&AoA of words per Sent &0.537&28   &0.502&2   &0.722&41   &0.711&6   &0.602&24\\
33&ShaTr&Shallow&as\_Chara\_C&\# characters per Sent &0.539&27   &0.487&4   &0.696&46   &0.711&5   &0.613&20\\
33&Synta&Tree Structure&as\_FTree\_C&length of flattened Trees per Sent &0.505&37   &0.485&5   &0.677&54   &0.719&2   &0.622&16\\
32&LxSem&Psycholinguistic&at\_AAKuL\_C&lemmas AoA of lemmas per Word &0.723&2   &0.323&35   &0.785&19   &0.65&20   &0.453&67\\
32&LxSem&Psycholinguistic&at\_AAKuW\_C&AoA of words per Word &0.703&5   &0.308&36   &0.784&20   &0.643&21   &0.455&66\\
\end{tabular}
}
\caption{\label{Table 8} Part A. The full generalizability ranking of handcrafted linguistic features under Approach A. r: Pearson's correlation between the feature and the dataset. rk: the feature's correlation ranking on the specific dataset.}
\end{table*}

\begin{table*}[t]
\centering
\resizebox{\textwidth}{!}{%
\begin{tabular}{
c@{\hspace{0.8ex}}
l@{\hspace{0.8ex}}
l@{\hspace{0.8ex}}
l@{\hspace{0.8ex}}
l@{\hspace{0.8ex}}
|c@{\hspace{0.8ex}}
c@{\hspace{0.8ex}}
|c@{\hspace{0.8ex}}
c@{\hspace{0.8ex}}
|c@{\hspace{0.8ex}}
c@{\hspace{0.8ex}}
|c@{\hspace{0.8ex}}
c@{\hspace{0.8ex}}
|c@{\hspace{0.8ex}}
c@{\hspace{0.8ex}}}
\cmidrule(lr){1-15}
\multicolumn{5}{c}{\textbf{Feature}} & \multicolumn{2}{c}{\textbf{CCB}} & \multicolumn{2}{c}{\textbf{WBT}} & \multicolumn{2}{c}{\textbf{CAM}} & \multicolumn{2}{c}{\textbf{CKC}} & \multicolumn{2}{c}{\textbf{OSE}} \\
\cmidrule(lr){1-5}\cmidrule(lr){6-7}\cmidrule(lr){8-9}\cmidrule(lr){10-11}\cmidrule(lr){12-13}\cmidrule(lr){14-15}
Score & Branch & Subgroup & LingFeat Code & Brief Explanation          &   r  &  rk &  r &  rk &  r  &  rk & r  &  rk& r  &   rk  \\
\cmidrule(lr){1-1}\cmidrule(lr){2-2}\cmidrule(lr){3-3}\cmidrule(lr){4-4}\cmidrule(lr){5-5}\cmidrule(lr){6-6}\cmidrule(lr){7-7}\cmidrule(lr){8-8}\cmidrule(lr){9-9}\cmidrule(lr){10-10}\cmidrule(lr){11-11}\cmidrule(lr){12-12}\cmidrule(lr){13-13}\cmidrule(lr){14-14}\cmidrule(lr){15-15}
25&ShaTr&Shallow&at\_Chara\_C&\# characters per Word &0.443&59   &0.2&52   &0.619&67   &0.402&83   &0.443&68\\
25&Synta&Phrasal&to\_SuPhr\_C&total \# Subordinate Clauses &0.367&96   &0.202&50   &0.721&43   &0.462&61   &0.419&78\\
25&LxSem&Psycholinguistic&to\_AABiL\_C&total lemmas AoA of lemmas, Bird norm &0.365&98   &0.155&79   &0.786&18   &0.473&59   &0.565&39\\
25&Synta&Part-of-Speech&to\_CoTag\_C&total \# Coordinating Conjunction POS tags &0.364&101   &0.268&43   &0.728&39   &0.406&80   &0.434&72\\
25&Synta&Part-of-Speech&to\_FuncW\_C&total \# Function words &0.33&126   &0.159&77   &0.773&25   &0.385&89   &0.636&14\\
25&Synta&Phrasal&to\_VePhr\_C&total \# Verb phrases &0.324&127   &0.169&68   &0.76&31   &0.416&72   &0.57&38\\
\end{tabular}
}
\caption{\label{Table 12} Part B. The full generalizability ranking of handcrafted linguistic features under Approach B.}
\end{table*}

\begin{table*}[t]
\centering
\resizebox{\textwidth}{!}{%
\begin{tabular}{
c@{\hspace{0.8ex}}
l@{\hspace{0.8ex}}
l@{\hspace{0.8ex}}
l@{\hspace{0.8ex}}
l@{\hspace{0.8ex}}
|c@{\hspace{0.8ex}}
c@{\hspace{0.8ex}}
|c@{\hspace{0.8ex}}
c@{\hspace{0.8ex}}
|c@{\hspace{0.8ex}}
c@{\hspace{0.8ex}}
|c@{\hspace{0.8ex}}
c@{\hspace{0.8ex}}
|c@{\hspace{0.8ex}}
c@{\hspace{0.8ex}}}
\cmidrule(lr){1-15}
\multicolumn{5}{c}{\textbf{Feature}} & \multicolumn{2}{c}{\textbf{CCB}} & \multicolumn{2}{c}{\textbf{WBT}} & \multicolumn{2}{c}{\textbf{CAM}} & \multicolumn{2}{c}{\textbf{CKC}} & \multicolumn{2}{c}{\textbf{OSE}} \\
\cmidrule(lr){1-5}\cmidrule(lr){6-7}\cmidrule(lr){8-9}\cmidrule(lr){10-11}\cmidrule(lr){12-13}\cmidrule(lr){14-15}
Score & Branch & Subgroup & LingFeat Code & Brief Explanation          &   r  &  rk &  r &  rk &  r  &  rk & r  &  rk& r  &   rk  \\
\cmidrule(lr){1-1}\cmidrule(lr){2-2}\cmidrule(lr){3-3}\cmidrule(lr){4-4}\cmidrule(lr){5-5}\cmidrule(lr){6-6}\cmidrule(lr){7-7}\cmidrule(lr){8-8}\cmidrule(lr){9-9}\cmidrule(lr){10-10}\cmidrule(lr){11-11}\cmidrule(lr){12-12}\cmidrule(lr){13-13}\cmidrule(lr){14-14}\cmidrule(lr){15-15}
12&LxSem&Variation Ratio&SimpNoV\_S&Noun Variation-1 &0.499&38   &0.087&130   &0.038&212   &0.031&213   &0.337&95\\
12&Synta&Part-of-Speech&ra\_NoVeT\_C&ratio of Noun POS \# to Verb POS \# &0.432&66   &0.118&111   &0.149&168   &0.112&170   &0.051&197\\
12&Synta&Phrasal&ra\_AvPrP\_C&ratio of Adv phrases \# to Prep phrases \# &0.4&82   &0.014&217   &0.222&147   &0.196&135   &0.115&152\\
12&Synta&Part-of-Speech&ra\_VeNoT\_C&ratio of Verb POS \# to Noun POS \# &0.397&86   &0.198&53   &0.234&142   &0.171&142   &0.067&186\\
12&Synta&Part-of-Speech&ra\_SuVeT\_C&ratio of Subordinating Conjunction POS \# to Verb POS \# &0.36&103   &0.053&159   &0.109&177   &0.282&115   &0.137&139\\
12&Disco&Entity Density&as\_UEnti\_C&number of unique Entities per Sent &0.337&121   &0.114&113   &0.273&136   &0.066&196   &0.157&128\\
\end{tabular}
}
\caption{\label{Table 13} Part C. The full generalizability ranking of handcrafted linguistic features under Approach B.}
\end{table*}

\begin{table*}[t]
\centering
\resizebox{\textwidth}{!}{%
\begin{tabular}{
c@{\hspace{0.8ex}}
l@{\hspace{0.8ex}}
l@{\hspace{0.8ex}}
l@{\hspace{0.8ex}}
l@{\hspace{0.8ex}}
|c@{\hspace{0.8ex}}
c@{\hspace{0.8ex}}
|c@{\hspace{0.8ex}}
c@{\hspace{0.8ex}}
|c@{\hspace{0.8ex}}
c@{\hspace{0.8ex}}
|c@{\hspace{0.8ex}}
c@{\hspace{0.8ex}}
|c@{\hspace{0.8ex}}
c@{\hspace{0.8ex}}}
\cmidrule(lr){1-15}
\multicolumn{5}{c}{\textbf{Feature}} & \multicolumn{2}{c}{\textbf{CCB}} & \multicolumn{2}{c}{\textbf{WBT}} & \multicolumn{2}{c}{\textbf{CAM}} & \multicolumn{2}{c}{\textbf{CKC}} & \multicolumn{2}{c}{\textbf{OSE}} \\
\cmidrule(lr){1-5}\cmidrule(lr){6-7}\cmidrule(lr){8-9}\cmidrule(lr){10-11}\cmidrule(lr){12-13}\cmidrule(lr){14-15}
Score & Branch & Subgroup & LingFeat Code & Brief Explanation          &   r  &  rk &  r &  rk &  r  &  rk & r  &  rk& r  &   rk  \\
\cmidrule(lr){1-1}\cmidrule(lr){2-2}\cmidrule(lr){3-3}\cmidrule(lr){4-4}\cmidrule(lr){5-5}\cmidrule(lr){6-6}\cmidrule(lr){7-7}\cmidrule(lr){8-8}\cmidrule(lr){9-9}\cmidrule(lr){10-10}\cmidrule(lr){11-11}\cmidrule(lr){12-12}\cmidrule(lr){13-13}\cmidrule(lr){14-14}\cmidrule(lr){15-15}
7&LxSem&Word Familiarity&at\_SbFrL\_C&SubtlexUS FREQlow value per Word &0.176&178   &0.171&65   &0.061&203   &0.001&228   &0.09&165\\
7&Synta&Part-of-Speech&ra\_AvNoT\_C&ratio of Adverb POS \# to Noun POS \# &0.171&179   &0.108&119   &0.076&195   &0.084&185   &0.023&222\\
7&Disco&Entity Grid&LoCoDPU\_S&Local Coherence distance for PU score &0.154&186   &0.032&189   &0.086&191   &0.087&182   &0.111&154\\
7&Synta&Phrasal&at\_AvPhr\_C&\# Adverb phrases per Word &0.144&188   &0.113&115   &0.047&208   &0.029&214   &0.058&191\\
7&Synta&Phrasal&ra\_AjSuP\_C&ratio of Adj phrases \# to Subordinate Clauses \# &0.133&194   &0.04&177   &0.195&155   &0.001&227   &0.079&173\\
7&Synta&Phrasal&ra\_AjVeP\_C&ratio of Adj phrases \# to Verb phrases \# &0.104&201   &0.01&222   &0.055&205   &0.083&186   &0.124&148\\
\end{tabular}
}
\caption{\label{Table 14} Part D. The full generalizability ranking of handcrafted linguistic features under Approach B.}
\vspace{-4mm}
\end{table*}

\end{document}